\documentclass[conference]{IEEEtran}
\IEEEoverridecommandlockouts
\usepackage{url}
\usepackage{cite}
\usepackage{amsmath,amssymb,amsfonts}
\usepackage{algorithmic}
\usepackage{graphicx}
\usepackage{textcomp}
\usepackage{xcolor}
\usepackage{booktabs}
\usepackage{bm}
\usepackage{hyperref}
\usepackage{cleveref}
\usepackage{multirow}
\def\BibTeX{{\rm B\kern-.05em{\sc i\kern-.025em b}\kern-.08em
    T\kern-.1667em\lower.7ex\hbox{E}\kern-.125emX}}
\begin{document}

\definecolor{myorangedark}{RGB}{201, 92, 46}
\definecolor{myreddark}{RGB}{151,79,75}
\definecolor{mybluedark}{RGB}{48,112,183}
\definecolor{mylightgreen}{RGB}{88,189,182}

\definecolor{plotpurple}{RGB}{207,45,219}
\definecolor{plotdarkblue}{RGB}{1,0,245}
\definecolor{plotblue}{RGB}{48,112,183}
\definecolor{plotorange}{RGB}{201,92,46}

\newcommand\blfootnote[1]{%
  \begingroup
  \renewcommand\thefootnote{}\footnote{#1}%
  \addtocounter{footnote}{-1}%
  \endgroup
}

\title{Comparing Self-Supervised Learning Models Pre-Trained on Human Speech and Animal Vocalizations for Bioacoustics Processing\\
\thanks{Source code: \href{https://github.com/idiap/ssl-human-animal}{\texttt{https://github.com/idiap/ssl-human-animal}}.}}

\author{\IEEEauthorblockN{Eklavya Sarkar\IEEEauthorrefmark{1}\IEEEauthorrefmark{2}, Mathew Magimai.-Doss\IEEEauthorrefmark{1}}
\IEEEauthorblockA{\IEEEauthorrefmark{1}Idiap Research Institute, Martigny, Switzerland}
\IEEEauthorblockA{\IEEEauthorrefmark{2}Ecole polytechnique f\'ed\'erale de Lausanne, Switzerland}
\texttt{\{eklavya.sarkar, mathew\}@idiap.ch}
}

\maketitle

\begin{abstract}
Self-supervised learning (SSL) foundation models have emerged as powerful, domain-agnostic, general-purpose feature extractors applicable to a wide range of tasks. Such models pre-trained on human speech have demonstrated high transferability for bioacoustic processing. This paper investigates (i) whether SSL models pre-trained directly on animal vocalizations offer a significant advantage over those pre-trained on speech, and (ii) whether fine-tuning speech-pretrained models on automatic speech recognition (ASR) tasks can enhance bioacoustic classification. We conduct a comparative analysis using three diverse bioacoustic datasets and two different bioacoustic tasks. Results indicate that pre-training on bioacoustic data provides only marginal improvements over speech-pretrained models, with comparable performance in most scenarios. Fine-tuning on ASR tasks yields mixed outcomes, suggesting that the general-purpose representations learned during SSL pre-training are already well-suited for bioacoustic tasks. These findings highlight the robustness of speech-pretrained SSL models for bioacoustics and imply that extensive fine-tuning may not be necessary for optimal performance.
\end{abstract}


\section{Introduction} \label{sec:intro}
Bioacoustics, the study of animal sounds, plays a crucial role in ecological and evolutionary research, providing insights into animal communication, biodiversity, and the origins of language. However, despite its significance, working with bioacoustic data presents several challenges: the data is often scarce, difficult to collect, noisy, and expensive to annotate. In recent years, advances in machine learning have made substantial progress in addressing these challenges \cite{bioacoustics_roadmap}. Notably, modern pre-trained deep learning foundation models have demonstrated impressive transferability to bioacoustic tasks, significantly advancing the field \cite{beans, Ghani2023, DUFOURQ2022101688, 10626094, 10627576}. In particular, self-supervised learning (SSL) models pre-trained on \textit{human speech} have shown remarkable success in tackling various bioacoustic tasks, such as animal call-type classification \cite{sarkar24_interspeech, mahoud24_interspeech, abzaliev24, kloots24_vihar, shi24_vihar}, caller identification \cite{sarkar23_interspeech, cauzinille24_interspeech, Knight2024}, and species recognition \cite{aves}. These models leverage large volumes of unlabeled data, prevalent in bioacoustics, by creating surrogate labels based on the intrinsic structure of the audio data, and then solving pre-text tasks designed to learn salient representations \cite{ssl_review}. Given the domain-agnostic nature of these pre-training tasks, SSL models have been effective in transferring from speech to bioacoustics without the need for domain-specific fine-tuning. Essentially, SSLs serve as powerful, general-purpose feature extractors for a wide range of downstream tasks.

Building on these developments, this paper explores the following two points, aimed at analyzing SSLs for bioacoustics:
\\
\textbf{1. SSL Pre-training Domain}: While SSL models pre-trained on human speech have shown strong transferability to bioacoustic tasks, recent research has explored pre-training on bioacoustic data itself, both in supervised and self-supervised frameworks \cite{birdnet, perch, aves}. The motivation behind pre-training on animal data is that these models may better capture species-specific vocal patterns and other properties unique to animal sounds. However, given that SSL pre-training is designed to learn general, domain-agnostic features, it is not yet clear whether pre-training directly on bioacoustics actually provides any significant advantage over SSLs pre-trained on human speech. Therefore, in this study, we systematically compare SSL models pre-trained on  human speech against those on animal vocalizations, and evaluate their performance for bioacoustics processing across various datasets and tasks.
\\
\textbf{2. Fine-tuning on Human Speech}: SSL representations have demonstrated strong performance on bioacoustic tasks without requiring fine-tuning, indicating their extracted latent representations can capture acoustically rich information capable of distinguishing animal call-types and caller identities. However, fine-tuning in a supervised framework often forces the model to learn novel and more specialized patterns, such as phonetic distinctions and temporal structures, typically leading to further performance gains. As both human speech and animal calls encode structured vocal and linguistic information for communication, SSL models fine-tuned on \textit{speech recognition} (ASR) may provide an additional inductive bias, enhancing the model's ability to recognize complex features in bioacoustic data. Therefore, we seek to explore whether fine-tuning pre-trained SSLs on human speech tasks, such as ASR, can further improve these models' capability to process animal vocalizations by capturing the subtle spectro-temporal characteristics present in animal calls, which may otherwise remain underrepresented in general SSL pre-training.

The rest of the paper is organized as follows: \cref{sec:exp_setup} provides the experimental setup for the studies in this paper, \cref{sec:exp} presents and thoroughly analyzes the experiments' comparative results. Finally, \cref{sec:conclusion} concludes the paper.

\section{Experimental Setup\label{sec:exp_setup}}
\subsection{Datasets, Tasks, and Protocols}
\begin{figure}[!htb]
  \centering
  \includegraphics[width=\linewidth]{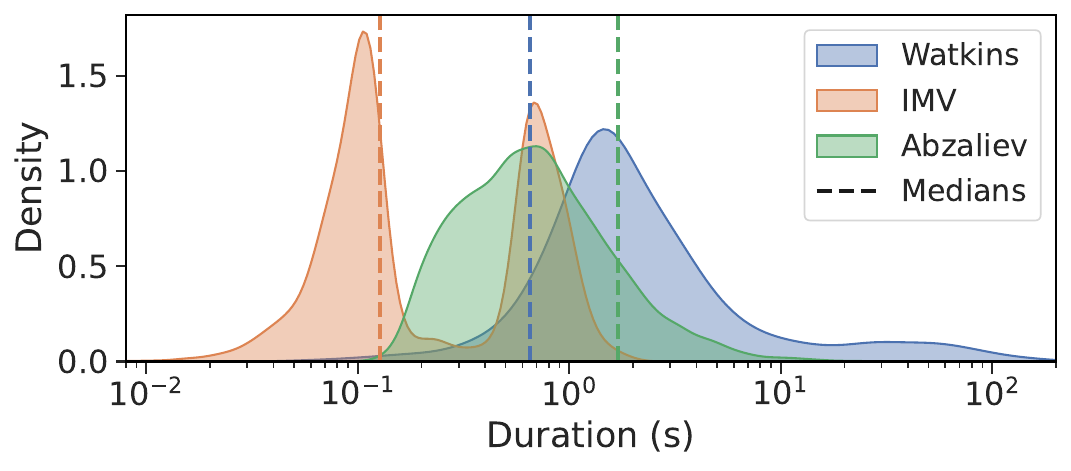}
  \vspace*{-0.8cm}
  \caption{Log distribution of vocalization lengths per dataset. The medians are calculated over the entirety of each dataset.}
  \label{fig:durations}
\end{figure}

We conducted the experiments for our studies on the three distinct bioacoustic datasets, summarized in \Cref{table:dataset_stats}. \Cref{fig:durations} also presents a log distribution of their vocalization lengths.
\begin{table}[!htb]
\centering
\caption{$L$ denotes the length [minutes], $n_c$ the number of classes, SR the sampling rate [kHz], $\mu$ the median length [ms], $\sigma$ the std.}
\begin{tabular}{lrrrrrr}
\toprule
\textbf{Dataset} & $\textbf{\# Samples}$ & $\bm{L}$ & \textbf{SR} & $\bm{n_c}$ & $\bm{\mu}$ & $\bm{\sigma}$ \\
\midrule
\cite{watkins} Watkins & $1,697$ & $295$ & -- & $32$ & $1701$ & $71245$ \\
\cite{sarkar23_interspeech} IMV & $72,920$ & $464$ & $44.1$ & $11$ & $127$ & $375$ \\
\cite{abzaliev24} Abzaliev & $8,034$ & $137$ & $48$ & $14$ & $655$ & $1313$ \\
\bottomrule
\end{tabular}
\label{table:dataset_stats}
\end{table}

\textbf{Watkins} \cite{watkins}: contains the recordings of different marine mammals, such as specific dolphins, whales, and seals. We chose Watkins for its multi-species vocalizations, rich acoustic variety, and high variance in segment lengths (\cref{fig:durations}). It has been commonly used for bioacoustic benchmarking, particularly for evaluating modern deep learning models \cite{aves, beans}. We chose the `best of' cut of the original dataset, a selected subset from the original 15,000 samples in total, deemed to be of higher sound quality and to contain less noise. The final dataset contains 1697 vocalization segments from 32 different species, totalling to 295 minutes, with a median length of 1701s. The sampling rate (SR) varies according to the recorded species. 

\textbf{InfantMarmosetsVox} (IMV) \cite{sarkar23_interspeech}: is an audio dataset of \textit{Callithrix jacchus}, a highly vocal new world primate. Marmosets were chosen for their complex social system, which allows them to encode vital information in their calls, such as identity, group affiliation, and dialect. They serve as surrogate models to understand the evolutionary origins of human vocal communication for neuro-biologists. The dataset consists of 72,920 segments representing 11 different call-types over 464 minutes. It was recorded from five pairs of infant marmoset twins, each recorded individually in sound-proofed rooms at 44.1 kHz SR, without communication with other marmoset pairs or the experimenters. The audio recordings were manually labeled by an experienced researcher. Although a large dataset by bioacoustics standards, each segment is predominantly short, with a median length of 127 ms. The spectral range of the calls is mostly centered around 7-8 kHz, although there is some information present above 16 kHz \cite{sarkar24_interspeech}.

\textbf{Abzaliev} \cite{abzaliev24}: is a novel dog dataset (here referred to
by the first author’s name) consisting of 8,034 vocalizations from the v2017 Mescalina Bark ID dataset \cite{10.3233/JIFS-169509}. It contains 14 different call-types, ranging from normal, aggressive, fearful, and playful barks at strangers (IDs 0--3), to vocalizations related to owner interaction (4--5) and non-stranger/non-play sounds (6). It also contains postive or negative whines (7--8) and growls (9--10), barks associated with sadness or anxiety (11), and excitement upon the owner's arrival home (12). The recordings originate from various dog breeds, including Chihuahuas, French Poodles, and Schnauzers. The data was recorded at 48 kHz SR from a microphone, and followed a protocol designed and validated by experts in animal behavior. The dog vocalizations were induced by exposing the dogs to different types of external stimuli, with the participation of the owner and/or experimenter. We discard all the segments labelled as non-dog sounds, such as TV, cars, and appliances.

For our experiments, we divide the datasets into a \textit{Train}, \textit{Val}, and \textit{Test} sets, following a random 70:20:10 split protocol. 

\subsection{Models and Feature Representations}
For our experiments, we select four different models to obtain our various feature representations $\mathcal{F}$. These consist of neural representations extracted through pre-trained (PT) models on animal vocalizations or human speech in a self-supervised learning framework, as well as their counterparts fine-tuned (PT+FT) in a supervised approach. The different features and their various properties are tabulated in \cref{table:models}.

\begin{table}[ht]
\centering
\caption{\# Parameters $P$ [M] and feature dimension $D$ of selected models. LS represents LibriSpeech and AS is AudioSet.}
\begin{tabular}{llrrrl}
\toprule
\bm{$\mathcal{F}$} & \textbf{Corpus} & $\bm{P}$ & $\bm{D}$ & \textbf{TL} & \textbf{Type} \\
\midrule
\cite{aves} AVES-Bio        & FSD, AS, Bio     & 94.68    & 768  & 12 & PT \\
\cite{hsu2021hubert} HuBERT & LS 960     & 94.68  & 768 & 12  & PT\\
\midrule
 \cite{baevski2020wav2vec} W2V2             & LS 960     & 95.04  & 768 & 12 & PT\\
\cite{baevski2020wav2vec} W2V2-100h        & LS 960     & 95.04  & 768 & 12 & PT+FT \\
\cite{baevski2020wav2vec} W2V2-960h        & LS 960     & 95.04  & 768 & 12 & PT+FT\\
\midrule
\cite{wavlm} WLM        & LS 960     & 94.38 & 768 & 12 & PT\\
\cite{wavlm} WLM-100h  & LS 960     & 94.38 & 768 & 12 & PT+FT\\
\bottomrule
\end{tabular}
\label{table:models}
\end{table}

\textbf{SSL pre-trained on animal vocalizations:}
We look at the AVES models family \cite{aves}, which are essentially the same as HuBERT models, but pre-trained on bioacoustics data instead of human speech. We select them based on their effectiveness on numerous bioacoustic classification and detection tasks, as well as the extensive benchmarking. Although this model performs well compared to traditional classifiers \cite{aves}, its performance has not been directly compared to a regular HuBERT model pre-trained on speech. The AVES set are pre-trained on combinations of publicly available audio datasets, namely FSD50K \cite{fonseca2021fsd50k}, AudioSet \cite{gemmeke2017audio}, and VGGSound \cite{vggsound}, instead of human speech. Specifically, we chose the \textit{Bio} model, which was pre-trained on a masked-prediction task on a total of 142K audio segments (360 hours) of the \texttt{animal} label in the AudioSet ontology (ID: \texttt{/m/0jbk}) and VGGSound class group. It's architecture is based on HuBERT's \textit{base} model, and contains 12 encoder transformer layers (TL).

\textbf{SSL pre-trained on human speech:}
In order to directly compare our performance against AVES-Bio, we select the HuBERT \textit{base} model, pre-trained on a masked-prediction task. In addition, we also look at the \textit{base} WavLM, denoted as WLM, based on its demonstrated effectiveness in animal call and caller classification \cite{sarkar23_interspeech, sarkar24_interspeech}, as well as its versatility in speech processing tasks as benchmarked on the SUPERB challenge \cite{yang21c_interspeech}. Finally, we also use 
the \textit{base} Wav2Vec2 model, denoted as W2V2, pre-trained on a constrastive task. All three models were pre-trained on the 960-hour Librispeech dataset.

\textbf{SSL pre-trained and fine-tuned on human speech:} For our second study, we assess the impact of fine-tuning on models pre-trained on human speech for bioacoustic tasks. To that end, we use WLM fine-tuned on 100 hours of Librispeech, and W2V2 fine-tuned on both 100 and 960 hours of Librispeech. All 3 models are fine-tuned on a ASR task\footnote{All fine-tuned models are obtained from Huggingface, namely from the \href{https://huggingface.co/facebook}{\texttt{facebook}}, \href{https://huggingface.co/microsoft}{\texttt{microsoft}}, and \href{https://huggingface.co/patrickvonplaten}{\texttt{patrickvonplaten}} repositories.}.

\textbf{Fusion}: We also compute a simple fusion representation as comparison to the other features. For each vocalization segment, we simply compute the mean across the posterior probabilities of all the other features, and then take its argmax.
\begin{figure}[!htb]
  \centering
  \includegraphics[width=\linewidth]{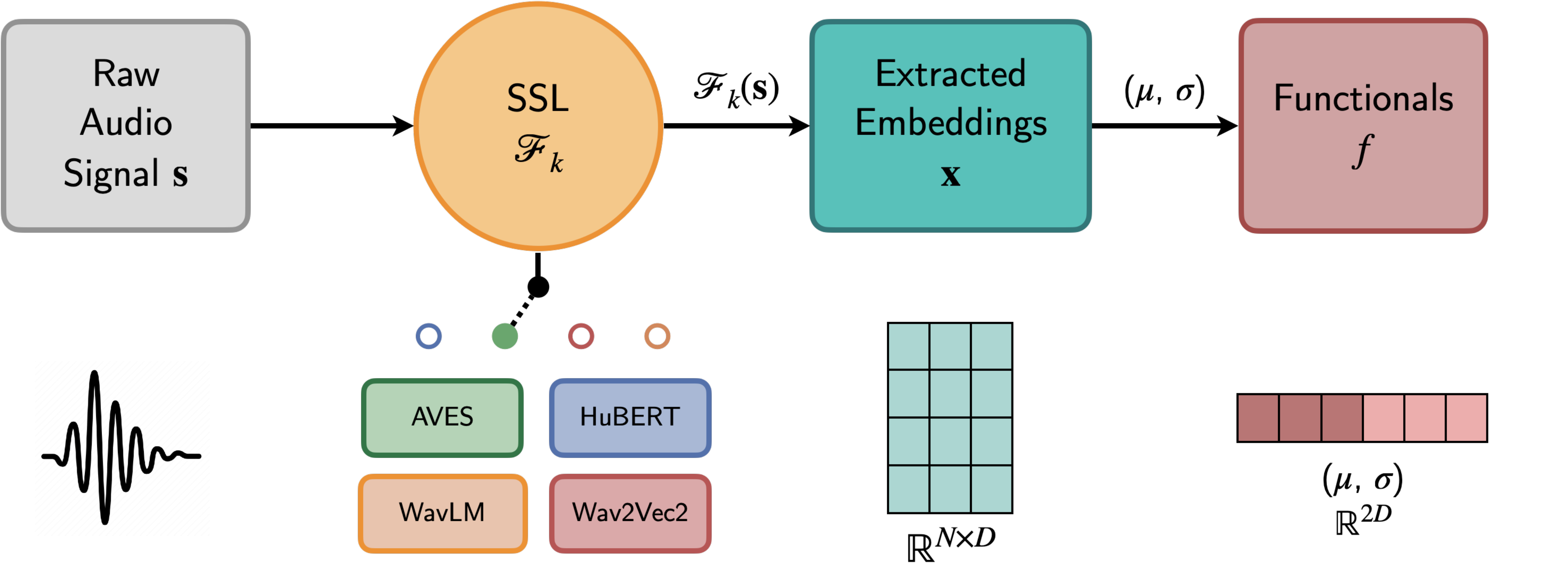}
  \vspace*{-0.8cm}
  \caption{Feature representation extraction pipeline.}
  \label{fig:pipeline}
\end{figure}

The general pipeline for obtaining a feature vector for a given vocalization segment is illustrated in \cref{fig:pipeline}. We obtain the features from these each of the SSL models $\mathcal{F}$, by first giving them the animal vocalizations $\bm{s}$ as inputs resampled at 16 kHz. We extract the variable-length embeddings $\bm{x} \in \mathbb{R}^{N \times D}$ output for each frame. Then, we transform them into fixed-length vocalization-level representations by computing and aggregating first and second order statistics across the temporal axis, resulting in a final feature functional representation $\bm{f} \in \mathbb{R}^{2D}$. For our work, we extract the embeddings of the CNN and all encoder transformer layers (TL) of $\mathcal{F}$, since we are interested in investigating the features at a layer level.

\section{Experiments and Analysis\label{sec:exp}}
This section looks at the classification performance of the extracted feature representations. In order to compare and evaluate the saliency of the different features, we follow existing literature \cite{sarkar24_interspeech} and classify them using a simple, non-linear MLP, composed of three blocks of [Linear, LayerNorm, ReLU] layers, with 128, 64, and 32 number of hidden units respectively, followed by a final linear layer.

We train the classifier for 30 epochs using cross-entropy loss, and employ a early-stopping criterion, where training is stopped if no improvement is observed on the \textit{Val} set for 10 consecutive epochs. The optimization consists of Adam, with a $\eta$-scheduler of factor 0.1 and patience of 10 epochs. We evaluate the performance through Unweighted Average Recall (UAR) as the metric to account for any class imbalance. 

\subsection{Pre-Training Domain Analysis}
\begin{figure}[h]
\vspace*{-0.3cm}
  \centering
  \includegraphics[width=\linewidth]{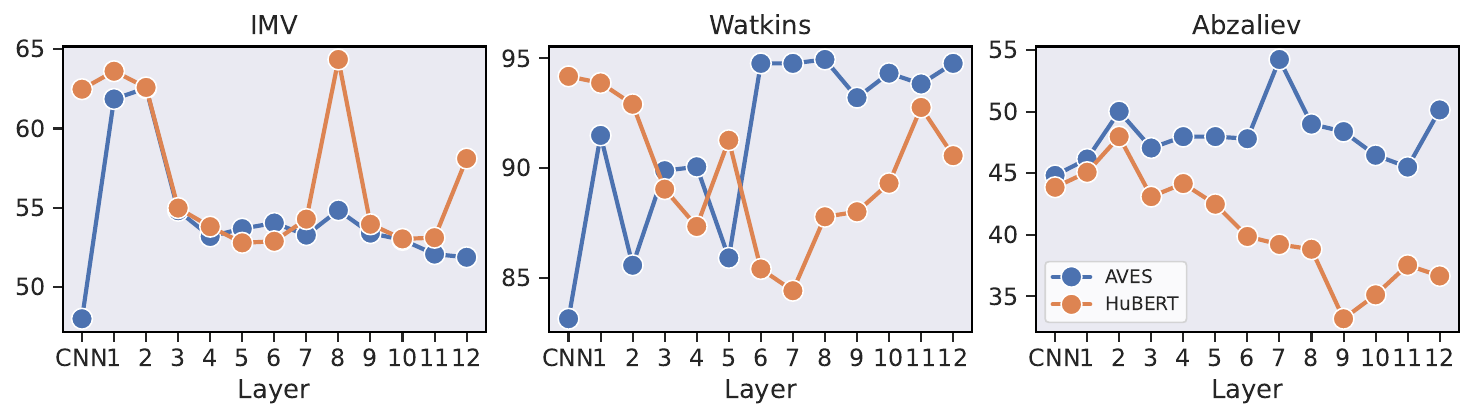}
  \vspace*{-0.8cm}
  \caption{Layer-wise performance of AVES (\textcolor{plotblue}{$\bullet$}) against HuBERT (\textcolor{plotorange}{$\bullet$}).}
  \label{fig:layers_domain}
\end{figure}

In this sub-section, we analyze the impact of pre-training domain by comparing AVES against HuBERT. \Cref{fig:layers_domain} shows that HuBERT outperforms AVES in the initial and final layers for IMV. Both models show that the initial transformer layers are more important for this task, indicating that this trend is not specific to speech-based pre-training. The loss of substantial spectral information in these Marmoset calls when down-sampled to 16 kHz likely affects the overall performance \cite{sarkar24_interspeech}. For Watkins, we see that AVES's initial layers are not as salient as later ones, where as HuBERT's middle layers are conversely the least useful. In the Abzaliev dataset, AVES performs better overall, with both the initial and later layers contributing comparably. HuBERT, on the other hand, does not scale well, and follows the same downwards trend as IMV. Overall, the results indicate that pre-training on bioacoustic data can provide marginal improvements in some datasets.

\subsection{Fine-Tuning Analysis}
\begin{figure}[h]
  \centering
  \includegraphics[width=\linewidth]{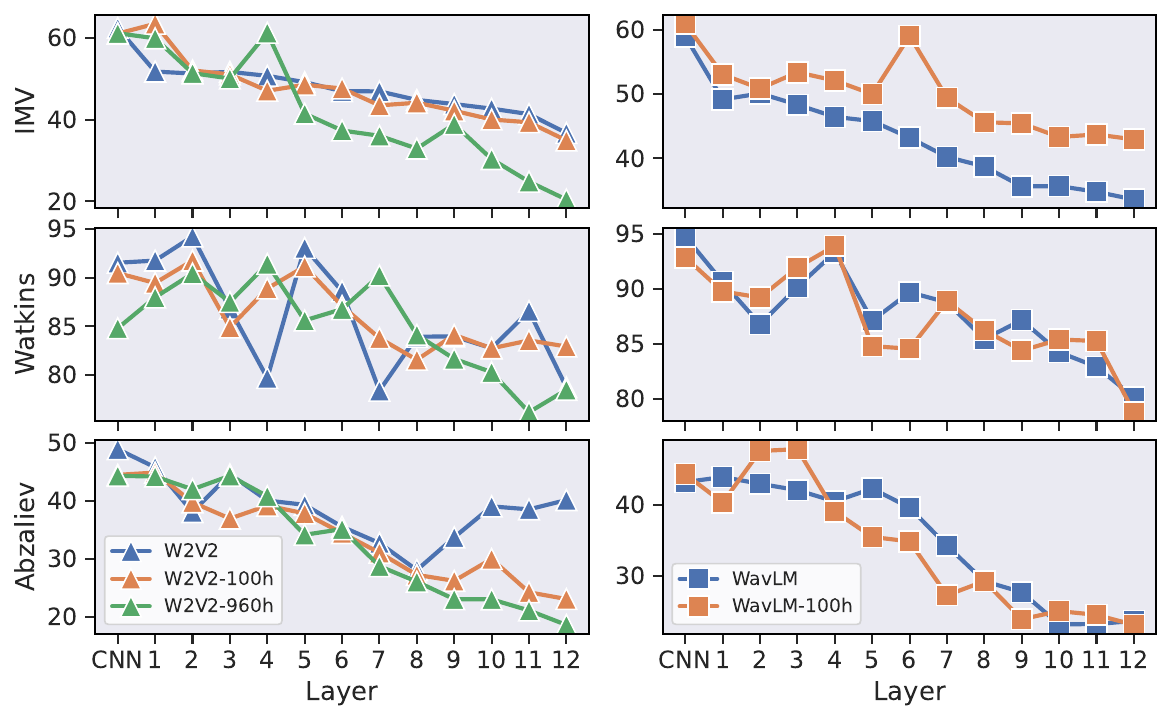}
  \vspace*{-0.8cm}
  \caption{UAR of W2V2 ({$\blacktriangle$}) and WLM ({$\blacksquare$}) against their fine-tuned versions.}
  \label{fig:layers_ft}
\end{figure}
Fine-tuning yields mixed effects across both models and datasets, as shown in \Cref{fig:layers_ft}. In several cases, we observe that fine-tuned models do not consistently outperform their base counterparts, particularly in W2V2-960h, with performance gains being marginal at best. Notably, fine-tuning on more speech data, such as the 960-hour W2V2, sometimes leads to a decline in performance in later layers, as seen on IMV and Abzaliev. This suggests that fine-tuning on speech may push models to learn task-specific features that don't generalize as well to certain bioacoustic tasks.

Interestingly, for non-fine-tuned models, earlier layers often capture enough general acoustic features to perform adequately. However, for fine-tuned models, selecting the optimal layer becomes more important, as different layers may capture more specialized representations that could benefit certain tasks. This points to the fact that fine-tuning creates more task-specific representations, making careful layer selection more necessary for specific bioacoustic tasks.

\subsection{Comparative Analysis}
Finally, we look at the general classification performance. \Cref{table:best_layer_results} tabulates the result of the layers yielding the highest scores from the different features. 
\begin{table}[h!]
\centering
\caption{UAR scores [\%] on the best feature layer, on \textit{Test}. \\Best performance is \textbf{bolded}, second best is \underline{underlined}.}
\begin{tabular}{clrrr}
\toprule
\textbf{Type} & \bm{$\mathcal{F}$} & \textbf{IMV} & \textbf{Watkins} & \textbf{Abzaliev} \\
\midrule
\multirow{4}{*}{PT} & AVES & 62.54 & \textbf{94.95} & \textbf{54.23} \\
& HuBERT & \textbf{64.35} & 94.18 & 47.96 \\
& WavLM & 58.98 & \underline{94.78} & 43.97 \\
& W2V2 & 62.40 & 94.25 & \underline{48.95} \\
\midrule
\multirow{3}{*}{PT + FT} & WavLM-100h & 60.93 & 93.93 & 47.90 \\
& W2V2-100h & \underline{63.44} & 91.77 & 44.91 \\
& W2V2-960h & 61.25 & 91.42 & 44.36 \\
\midrule
& Fusion & 62.48 & 94.78 & 48.95 \\
\bottomrule
\end{tabular}
\label{table:best_layer_results}
\end{table}

We can observe that the best scores are from the AVES and HuBERT models, both of which consist of the same architecture, pre-text task, and loss function. HuBERT and AVES yield very comparable performances for both IMV and Watkins, indicating that HuBERT's representations are robust for call-type classification tasks across different species. AVES achieves a higher score on the Watkins dataset, suggesting that for this specific task, pre-training on bioacoustic data yields a small but notable improvement for species classification. Additionally, we can clearly observe that all the best scores are from the PT category, as well as the second best scores with the marginal exception W2V2-100h on the IMV dataset. This demonstrates that further fine-tuning pre-trained speech models on an ASR task does not consistently bring us any advantage over the pre-trained alone for bioacoustics classification tasks. It suggests that the pre-trained representations may already be optimized, and fine-tuning might not always yield significant benefits. Lastly, we observe that a fusion of all features over their best layers doesn't yield a more salient representation than the best performing model, although it can outperform some of the others.
\begin{figure}[h!]
  \centering
  \includegraphics[width=\linewidth]{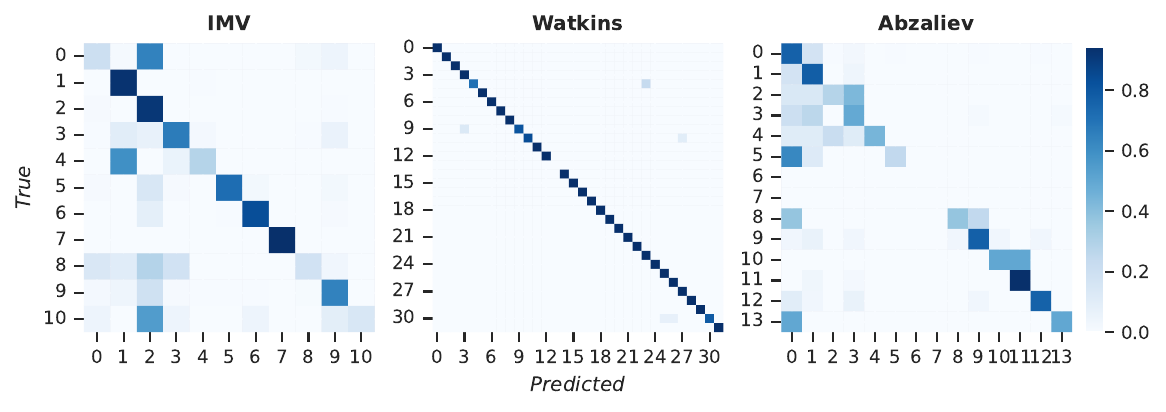}
  \vspace*{-0.8cm}
  \caption{Confusion matrices of the best feature layers' fusion.}
  \label{fig:cms_fusion}
\end{figure}

\Cref{fig:cms_fusion} shows the classifier's performance of the fusion features through confusion matrices. We can observe a good classification alignment for the three datasets. For IMV, there is a noticeable trend of false positives for call-type ID 2, likely due to its high occurrence in the dataset, and wide spectral range, causing an overlap of acoustic features with the other classes. The Watkins dataset is unsurprisingly the easiest to classify, likely because of the clear acoustic and spectral differences in the various species vocalizations, as well as the high variance in segment lengths. Class ID 13 only had two samples which results in an empty row. For Abzaliev, we observe some confusion between the different barks (IDs 0--5) which may contain overlapping acoustic features. Some classes had very few samples (ID 6), or were removed during data preprocessing (ID 7), resulting in empty rows.

\section{Conclusion\label{sec:conclusion}}
This paper presented a comparison of self-supervised learning models pre-trained on human speech and animal vocalizations for bioacoustic tasks. Through two distinct lines of investigation, we first examined the impact of pre-training domains by comparing models pre-trained on human speech and animal vocalizations. The results indicated that pre-training on bioacoustic data mostly yields comparable performance to pre-training on speech, but can offer limited advantages in select contexts. In our second line of investigation, we explored whether fine-tuning pre-trained speech models on ASR could further enhance their ability to capture structured patterns in animal vocalizations. We found that fine-tuning yielded inconsistent results, suggesting that the general-purpose representations learned during pre-training may already be well-suited for bioacoustic tasks, and further fine-tuning on speech does not consistently provide additional benefits.

In conclusion, our results highlight the utility of pre-trained speech models for bioacoustic tasks, even without further fine-tuning. Future work could explore attention mechanisms in SSL models to gain deeper insights into how these models interpret and process specific features of animal vocalizations.

\section*{Acknowledgment}
This work was funded by Swiss National Science Foundation's NCCR Evolving Language (grant no. 51NF40\_180888). The Mescalina Bark ID Data Base is a property of Efecto Mescalina SAPI de CV. We would like to thank Dr. Humberto P\'erez-Espinosa, for their dedication of the collection of the Mescalina Bark ID data base. 


\clearpage
\bibliographystyle{IEEEtran}
\bibliography{refs}

\begin{thebibliography}{10}
\providecommand{\url}[1]{#1}
\csname url@samestyle\endcsname
\providecommand{\newblock}{\relax}
\providecommand{\bibinfo}[2]{#2}
\providecommand{\BIBentrySTDinterwordspacing}{\spaceskip=0pt\relax}
\providecommand{\BIBentryALTinterwordstretchfactor}{4}
\providecommand{\BIBentryALTinterwordspacing}{\spaceskip=\fontdimen2\font plus
\BIBentryALTinterwordstretchfactor\fontdimen3\font minus \fontdimen4\font\relax}
\providecommand{\BIBforeignlanguage}[2]{{%
\expandafter\ifx\csname l@#1\endcsname\relax
\typeout{** WARNING: IEEEtran.bst: No hyphenation pattern has been}%
\typeout{** loaded for the language `#1'. Using the pattern for}%
\typeout{** the default language instead.}%
\else
\language=\csname l@#1\endcsname
\fi
#2}}
\providecommand{\BIBdecl}{\relax}
\BIBdecl

\bibitem{bioacoustics_roadmap}
D.~Stowell, ``{Computational bioacoustics with deep learning: a review and roadmap},'' \emph{PeerJ}, vol.~10, p. e13152, 2022.

\bibitem{beans}
M.~Hagiwara, B.~Hoffman, J.-Y. Liu, M.~Cusimano, F.~Effenberger, and K.~Zacarian, ``Beans: The benchmark of animal sounds,'' in \emph{Proc. of ICASSP}, 2023, pp. 1--5.

\bibitem{Ghani2023}
B.~Ghani, T.~Denton, S.~Kahl, and H.~Klinck, ``Global birdsong embeddings enable superior transfer learning for bioacoustic classification,'' \emph{Scientific Reports}, vol.~13, no.~1, p. 22876, 2023.

\bibitem{DUFOURQ2022101688}
E.~Dufourq, C.~Batist, R.~Foquet, and I.~Durbach, ``Passive acoustic monitoring of animal populations with transfer learning,'' \emph{Ecological Informatics}, vol.~70, p. 101688, 2022.

\bibitem{10626094}
C.~Heggan, S.~Budgett, T.~Hospedales, and M.~Yaghoobi, ``On the transferability of large-scale self-supervision to few-shot audio classification,'' in \emph{2024 IEEE International Conference on Acoustics, Speech, and Signal Processing Workshops (ICASSPW)}, 2024, pp. 515--519.

\bibitem{10627576}
I.~Moummad, N.~Farrugia, and R.~Serizel, ``Self-supervised learning for few-shot bird sound classification,'' in \emph{2024 IEEE International Conference on Acoustics, Speech, and Signal Processing Workshops (ICASSPW)}, 2024, pp. 600--604.

\bibitem{sarkar24_interspeech}
E.~Sarkar and M.~Magimai-Doss, ``{On the Utility of Speech and Audio Foundation Models for Marmoset Call Analysis},'' in \emph{Proc. of 4th International Workshop on Vocal Interactivity in-and-between Humans, Animals and Robots (VIHAR)}, 2024.

\bibitem{mahoud24_interspeech}
I.~B. Mahoud, E.~Sarkar, and M.~Magimai.-Doss, ``{Feature Representations for Automatic Meerkat Vocalization Classification},'' in \emph{Proc. of 4th International Workshop on Vocal Interactivity in-and-between Humans, Animals and Robots (VIHAR)}, 2024.

\bibitem{abzaliev24}
A.~Abzaliev, H.~Perez-Espinosa, and R.~Mihalcea, ``Towards dog bark decoding: Leveraging human speech processing for automated bark classification,'' in \emph{Proceedings of the 2024 Joint International Conference on Computational Linguistics, Language Resources and Evaluation (LREC-COLING 2024)}, N.~Calzolari, M.-Y. Kan, V.~Hoste, A.~Lenci, S.~Sakti, and N.~Xue, Eds.\hskip 1em plus 0.5em minus 0.4em\relax Torino, Italia: ELRA and ICCL, May 2024.

\bibitem{kloots24_vihar}
M.~de~Heer~Kloots and M.~Knornschild, ``{Exploring bat song syllable representations in self-supervised audio encoders},'' in \emph{Proc. of 4th International Workshop on Vocal Interactivity in-and-between Humans, Animals and Robots (VIHAR)}, 2024.

\bibitem{shi24_vihar}
R.~Shi, K.~Itoyama, and K.~Nakadai1, ``{Bird Vocalization Embedding Extraction Using Self-Supervised Disentangled Representation Learning},'' in \emph{Proc. of 4th International Workshop on Vocal Interactivity in-and-between Humans, Animals and Robots (VIHAR)}, 2024.

\bibitem{sarkar23_interspeech}
E.~Sarkar and M.~Magimai.-Doss, ``{Can Self-Supervised Neural Representations Pre-Trained on Human Speech distinguish Animal Callers?}'' in \emph{Proc. of Interspeech}, 2023, pp. 1189--1193.

\bibitem{cauzinille24_interspeech}
J.~Cauzinille, B.~Favre, R.~Marxer, D.~Clink, A.~H. Ahmad, and A.~Rey, ``Investigating self-supervised speech models' ability to classify animal vocalizations: The case of gibbon's vocal signatures,'' in \emph{Proc. of Interspeech}, 2024.

\bibitem{Knight2024}
E.~Knight, T.~Rhinehart, D.~R. de~Zwaan, M.~J. Weldy, M.~Cartwright, S.~H. Hawley, J.~L. Larkin, D.~Lesmeister, E.~Bayne, and J.~Kitzes, ``Individual identification in acoustic recordings,'' \emph{Trends in Ecology \& Evolution}, 2024.

\bibitem{aves}
M.~Hagiwara, ``Aves: Animal vocalization encoder based on self-supervision,'' in \emph{Proc. of ICASSP}, 2023, pp. 1--5.

\bibitem{ssl_review}
A.~Mohamed, H.-y. Lee, L.~Borgholt, J.~D. Havtorn, J.~Edin, C.~Igel, K.~Kirchhoff, S.-W. Li, K.~Livescu, L.~Maaløe, T.~N. Sainath, and S.~Watanabe, ``Self-supervised speech representation learning: A review,'' \emph{IEEE Journal of Selected Topics in Signal Processing}, vol.~16, no.~6, pp. 1179--1210, 2022.

\bibitem{birdnet}
S.~Kahl, C.~M. Wood, M.~Eibl, and H.~Klinck, ``Birdnet: A deep learning solution for avian diversity monitoring,'' \emph{Ecological Informatics}, vol.~61, p. 101236, 2021.

\bibitem{perch}
G.~Research, ``Perch,'' \url{https://github.com/google-research/perch}, (Accessed on 09/12/2024).

\bibitem{watkins}
L.~Sayigh, M.~A. Daher, J.~Allen, H.~Gordon, K.~Joyce, C.~Stuhlmann, and P.~Tyack, ``{The Watkins Marine Mammal Sound Database: An online, freely accessible resource},'' \emph{Proceedings of Meetings on Acoustics}, vol.~27, no.~1, p. 040013, 02 2017.

\bibitem{10.3233/JIFS-169509}
H.~P\'{e}rez-Espinosa, V.~Reyes-Meza, E.~Aguilar-Benitez, Y.~M. Sanz\'{o}n-Rosas, D.~Pinto, V.~K. Singh, A.~Villavicencio, P.~Mayr-Schlegel, and E.~Stamatatos, ``Automatic individual dog recognition based on the acoustic properties of its barks,'' \emph{Journal of Intelligent \& Fuzzy Systems}, vol.~34, no.~5, p. 3273–3280, Jan. 2018.

\bibitem{hsu2021hubert}
W.-N. Hsu, B.~Bolte, Y.-H.~H. Tsai, K.~Lakhotia, R.~Salakhutdinov, and A.~Mohamed, ``{HuBERT: Self-supervised speech representation learning by masked prediction of hidden units},'' \emph{IEEE/ACM Transactions on Audio, Speech, and Language Processing}, vol.~29, pp. 3451--3460, 2021.

\bibitem{baevski2020wav2vec}
A.~Baevski, Y.~Zhou, A.~Mohamed, and M.~Auli, ``{wav2vec 2.0: A framework for self-supervised learning of speech representations},'' \emph{Advances in Neural Information Processing Systems}, vol.~33, pp. 12\,449--12\,460, 2020.

\bibitem{wavlm}
S.~C. et~al., ``{WavLM: Large-Scale Self-Supervised Pre-Training for Full Stack Speech Processing},'' \emph{IEEE Journal of Selected Topics in Signal Processing}, vol.~16, no.~6, 2022.

\bibitem{fonseca2021fsd50k}
E.~Fonseca, X.~Favory, J.~Pons, F.~Font, and X.~Serra, ``Fsd50k: an open dataset of human-labeled sound events,'' \emph{IEEE/ACM Transactions on Audio, Speech, and Language Processing}, vol.~30, 2021.

\bibitem{gemmeke2017audio}
J.~F. Gemmeke, D.~P. Ellis, D.~Freedman, A.~Jansen, W.~Lawrence, R.~C. Moore, M.~Plakal, and M.~Ritter, ``Audio set: An ontology and human-labeled dataset for audio events,'' in \emph{Proc. of ICASSP}, 2017.

\bibitem{vggsound}
H.~Chen, W.~Xie, A.~Vedaldi, and A.~Zisserman, ``Vggsound: A large-scale audio-visual dataset,'' in \emph{Proc. of ICASSP}, 2020, pp. 721--725.

\bibitem{yang21c_interspeech}
S.~wen Yang~et al., ``{SUPERB: Speech Processing Universal PERformance Benchmark},'' in \emph{Proc. Interspeech}, 2021, pp. 1194--1198.

\end{thebibliography}

\end{document}